\documentclass[runningheads]{llncs}

\usepackage{eccv}

\usepackage{eccvabbrv}

\usepackage{graphicx}
\usepackage{booktabs}
\usepackage{colortbl}
\definecolor{tabfirst}{rgb}{1, 0.7, 0.7}
\definecolor{tabsecond}{rgb}{1, 0.85, 0.7}
\definecolor{tabthird}{rgb}{1, 1, 0.7}

\usepackage[accsupp]{axessibility}

\usepackage{algorithm}
\usepackage{algpseudocode}
\algtext*{EndFor}
\usepackage{stfloats}
\usepackage[hypcap=false]{caption}
\usepackage{multirow}
\usepackage{wrapfig}
\usepackage{pifont}
\usepackage{enumitem}
\usepackage{float}
\usepackage[utf8]{inputenc}
\usepackage{tcolorbox}
\tcbuselibrary{listings, skins, breakable}

\usepackage{hyperref}

\usepackage{orcidlink}

\begin{document}

\title{\texorpdfstring{Locality-Aware Continual Unlearning \\ for Diffusion Models}{Locality-Aware Continual Unlearning for Diffusion Models}}

\author{Naveen George\inst{1,2}$^\dagger$ \and
Naoki Murata\inst{2} \and
Yuhta Takida\inst{2} \and
Konda Reddy Mopuri\inst{1} \and
Yuki Mitsufuji\inst{2,3}}

\authorrunning{N.~George et al.}

\institute{Indian Institute of Technology Hyderabad, India \and
Sony AI, Japan \and Sony Group Corporation, Japan \\
\email{ai23mtech12001@iith.ac.in,naoki.murata@sony.com}
}

\maketitle
\begingroup
\renewcommand\thefootnote{\fnsymbol{footnote}}
\footnotetext[4]{Work done during an internship at Sony AI.}
\endgroup

\begin{abstract}
  Real-world deployment of text-to-image diffusion models requires continual concept removal as new privacy, copyright, or safety obligations arise over time. Existing unlearning methods, however, are designed for single-step deletion and collapse after only 3--5 sequential applications. We trace this instability to two compounding factors: \emph{(i)} coarse mapping targets that cause degradation to accumulate unnecessarily across steps, and \emph{(ii)} the absence of local protection for semantically neighboring concepts, whose shared internal representations make them the first to suffer collateral damage. Because this damage is strongest in the \emph{local} semantic neighborhood of the forget concept, global replay alone cannot prevent it. Building on this analysis, we propose \textbf{Locality-Aware Continual Unlearning (LACU)}, a framework with two complementary mechanisms. \textbf{Locality-Aware Target Selection} chooses, for each forget prompt, the context-preserving mapping prompt that the diffusion model itself treats as most similar to the original prompt, measured by \emph{score-prediction distance} (how differently the model denoises the same noisy image under two text conditions), ensuring each update is as small and targeted as possible. \textbf{Locality-Aware Replay} uses the same metric to identify the retain concepts closest to the forget concept in the model's own representation and replays them as a local functional regularizer, directly shielding the most vulnerable neighborhood. Combined with teacher-student distillation and lightweight $\ell_2$ parameter regularization, LACU maintains stable unlearning over 10 sequential steps, preserving significantly higher related retention ($RR_{\text{acc}}$) and general retention ($GR_{\text{acc}}$) than recent baselines. The code is available at \url{https://github.com/SonyResearch/LACU}.

\end{abstract}
    
\section{Introduction}
\label{sec:intro}

Recent visual generative models such as Sora~\cite{brooks2024video}, Gemini~\cite{team2023gemini}, Imagen 3.0~\cite{baldridge2024imagen}, and DALL·E 3~\cite{betker2023improving} synthesize high-quality content at scale.
Once deployed, these models need to accommodate deletion requests driven by privacy regulations~\cite{regulation2018general}, copyright claims, and safety policies.
Because such requests arrive continuously rather than as a one-time batch, \emph{continual unlearning} (CUL), i.e., sequentially removing concepts while preserving the remaining capabilities, is the operational baseline, not a special case.

\begin{figure*}[!t]
    \centering
    \includegraphics[width=\textwidth]{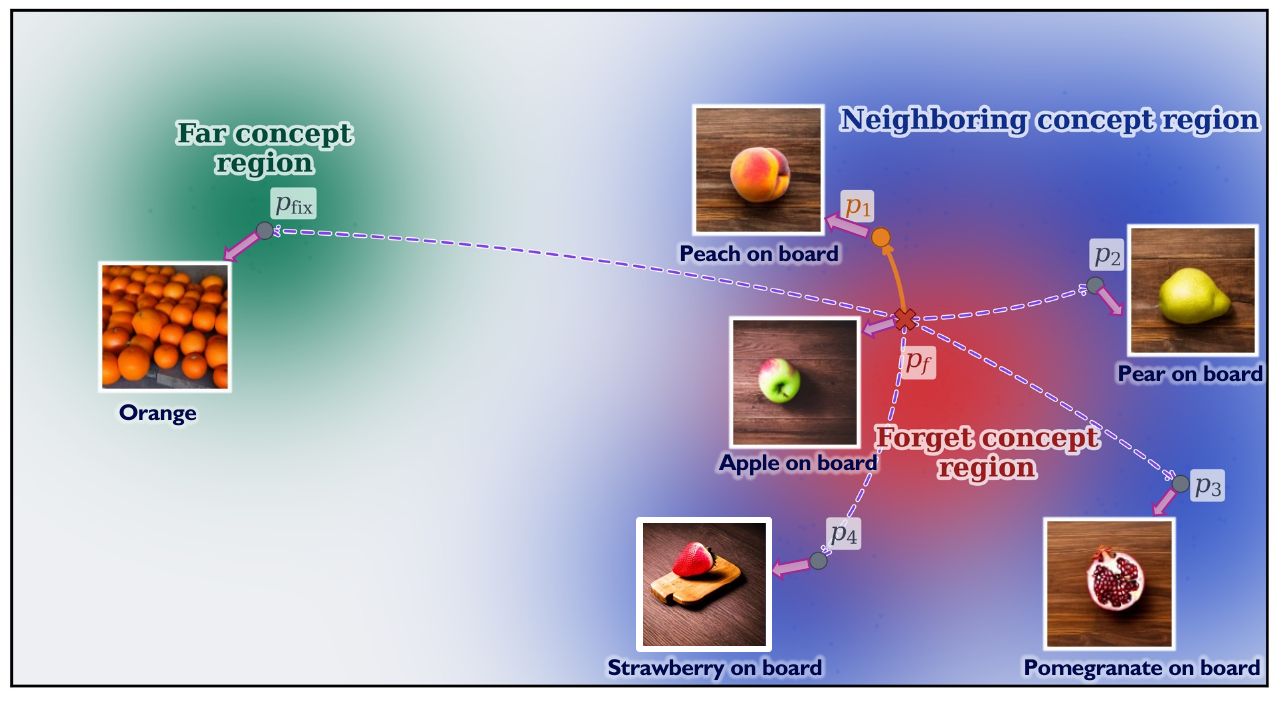}
    \captionof{figure}{\textbf{Locality-Aware Target Selection in the model's score-prediction space.}
    The landscape visualizes how the diffusion model itself perceives different prompts, where proximity is measured by \emph{score-prediction distance} $d_{\text{score}}$ (Eq.~\ref{eq:score_dist}): prompts the model denoises similarly lie close together.
    \textcolor{red}{\textbf{Red}}: forget concept; \textcolor{blue}{\textbf{blue}}: neighboring concepts; \textcolor{teal}{\textbf{teal}}: far concept.
    Given a forget prompt \textcolor{red}{$p_f$} (\textcolor{red}{``Apple on board''~\ding{55}}), our method generates context-preserving candidates $p_1$--$p_4$ that replace only the forget concept while retaining the scene context, and selects \textcolor{orange}{$p_1$} (\textcolor{orange}{``Peach on board''}) as the mapping target because it has the smallest $d_{\text{score}}$ to $p_f$.
    This proximity ensures only a small perturbation of the learned score field, minimizing collateral damage to neighbors.
    In contrast, traditional methods map all forget prompts to a fixed anchor $p_{\text{fix}}$ (``Orange'') that discards scene context, forcing a large displacement that disproportionately degrades neighboring concepts, compounding under sequential unlearning.}
    \label{fig:mapping_idea}
\end{figure*}

Current unlearning methods target a single, one-shot deletion and degrade rapidly under repeated application: as we show empirically (Section~\ref{sec:experiments}), most baselines collapse after only 3--5 sequential steps.
We hypothesize that the root cause is threefold.
(1)~Common mapping strategies redirect all forget prompts to the same fixed anchor (e.g., an empty string or a generic concept), which poorly approximates the ideal per-prompt mapping: different prompts containing the same forget concept can differ substantially in scene context, so a single surrogate forces unnecessarily \emph{large score-prediction displacements} at each step.
(2)~These methods lack \emph{local protection} for semantically neighboring concepts.
Prior benchmarks~\cite{zhang2024unlearncanvas,amara2025erasebench} show that the strongest collateral damage, called the \emph{ripple effect}, falls on concepts that are semantically close to the forget concept because these concepts share similar internal representations within the model.
Each oversized update compounds this local distortion; after several steps, unprotected neighbors degrade into \emph{retention collapse}.
Global replay (i.e., replaying random general prompts without targeting the local neighborhood) or generic regularization cannot adequately defend these vulnerable neighbors: what is needed is a protection mechanism that focuses on the \emph{local} semantic neighborhood of each forget concept.
(3)~Methods that do attempt more careful selection often rely on text-space similarity (e.g., CLIP embeddings), which does not necessarily reflect how the diffusion model itself processes different prompts; this mismatch further compounds errors under sequential application.

To address these issues, we propose \textbf{Locality-Aware Continual Unlearning (LACU)}, a framework whose two core components address locality through a single model-aware principle: measuring concept proximity relative to the forget target in the model's prediction space.

\begin{enumerate}
    \item \textbf{Locality-Aware Target Selection.}
    For each forget prompt, we generate a set of context-preserving candidate replacements and select the one whose noise predictions are closest to those of the forget prompt on shared noisy states.
    By choosing the \emph{nearest safe target}, each unlearning update becomes as small and precise as possible, reducing cumulative drift and limiting ripple effects on neighboring concepts.

    \item \textbf{Locality-Aware Replay.}
    Using the same score-prediction-distance-based metric, we identify the retain concepts that are \emph{closest to the forget concept in the model's own representation} and replay them as a local functional regularizer.
    This directly shields the most vulnerable neighborhood, precisely the concepts that benchmarks identify as first to break.

    \item \textbf{Stable 10-step continual unlearning.}
    Combined with teacher-student distillation and lightweight $\ell_2$ parameter regularization, LACU maintains stable performance across 10 sequential unlearning steps, preserving significantly higher related retention ($RR_{\text{acc}}$) and general retention ($GR_{\text{acc}}$) than recent baselines while effectively removing target concepts.
\end{enumerate}

\section{Related Work}
\label{sec:related}

\noindent\textbf{One-Shot Unlearning in Diffusion Models.}
Machine unlearning has become a key post-hoc mechanism for aligning generative models with evolving safety, privacy, and copyright requirements without full retraining.
Most existing approaches are single-step interventions that map forget concepts to fixed anchors.
ESD~\cite{gandikota2023erasing} steers away from the target using classifier-free guidance; Concept Ablation~\cite{kumari2023ablating} minimizes KL divergence toward a null anchor.
More parameter-efficient variants target cross-attention layers ~\cite{gandikota2024unified,zhang2024forget}, use efficient concept erasure~\cite{gong2024reliable}, train lightweight adapters~\cite{lyu2024one,huang2024receler,lu2024mace}, restrict updates via saliency~\cite{fan2024salun}, timestep selection~\cite{wu2024scissorhands}, or reduce forget--retain gradient conflicts~\cite{patel2025learning}.
Complementary work analyzes gradient-ascent failures, whether erased concepts are truly removed, and how to make erasure more reliable or adversarially robust~\cite{mavrothalassitis2026ascent,lu2025concepts,chen2025trce,srivatsan2025stereo}.
Despite their variety, these methods share a common weakness under repetition: \emph{coarse replacement} strategies map diverse forget prompts to the same fixed anchor, forcing score-prediction displacements that accumulate and destabilize the model~\cite{george2025illusion}.

\noindent\textbf{Continual Unlearning.}
Real deployments involve sequential removal requests, making CUL the practical setting.
Recent CUL analyses identify knowledge erosion, forgetting reversal, and cumulative parameter drift as key sequential failure modes~\cite{wuerkaixi2025adaptive,heng2023selective,thakral2025continual,lee2026continual,park2026robust,adhikari2025unlearning}.
DUGE~\cite{thakral2025continual} and regularization-based methods~\cite{lee2026continual} show that replay, distillation, and parameter regularization are useful backbones, but they mainly protect globally or in parameter space.

\noindent\textbf{Target Selection and Mapping.}
The choice of \emph{where to redirect} the forget concept substantially influences the magnitude of the resulting update.
AGE~\cite{bui2025fantastic} improves over fixed generic anchors by selecting an adaptive concept-level target from CLIP-filtered class labels through minimax optimization, supporting the view that forget concepts should be redirected to nearby non-synonym targets.
However, its targets remain discrete labels chosen outside the current diffusion score field, and the update does not explicitly protect neighboring retain concepts where ripple effects concentrate.
LACU addresses both issues by ranking prompt-level, context-preserving targets with score-prediction distance at each continual step and coupling the same locality measure with replay; AGE-style selection alone still yields high erasure but severe retention collapse in CUL (Section~\ref{sec:experiments}).

\noindent\textbf{Replay and Distillation for Retention.}
Knowledge distillation from a frozen teacher to a student model is a proven retention mechanism in both continual learning~\cite{hinton2015distilling,pmlr-v274-masip25a} and unlearning~\cite{chen2025score}.
Existing replay strategies typically sample from a broad, global prompt pool, which distributes the regularization signal uniformly.
However, prior benchmarks show that ripple effects concentrate in the \emph{local neighborhood} of the forget concept~\cite{zhang2024unlearncanvas,amara2025erasebench}; global replay therefore over-invests on already-safe regions and under-invests on the most fragile ones.
A replay strategy that explicitly prioritizes the nearest neighbors of the forget concept can therefore provide stronger protection where it matters most.

\section{Problem Formulation and Challenges}

\subsection{Continual Unlearning: Problem Setting}

We consider continual unlearning for text-to-image diffusion models in the setting where \emph{one forget concept is processed per step}. Let $\epsilon_{\theta_0}$ be a pre-trained model and let $\{c_{\text{f}}^{(1)}, c_{\text{f}}^{(2)}, \ldots, c_{\text{f}}^{(K)}\}$ be a sequence of forget concepts arriving over time. The model is updated sequentially as $\theta_i = \text{Unlearn}(\theta_{i-1}, c_{\text{f}}^{(i)})$ for $i \in \{1,\ldots,K\}$, while preserving a retain concept set $C_{\text{r}}$ such that $c_{\text{f}}^{(i)} \notin C_{\text{r}}$ for all $i$. At each step, the update must simultaneously satisfy three requirements:
\begin{enumerate}
    \item[(i)] \textit{Forgetting}: prompts instantiating $c_{\text{f}}^{(i)}$ should no longer produce recognizable target content.
    \item[(ii)] \textit{Retention}: non-target prompts remain behaviorally close to the previous model.
    \item[(iii)] \textit{Quality preservation}: overall image quality remains comparable to the original model.
\end{enumerate}
These three axes define the design space for subsequent sections: each component of our method is motivated by which requirement it primarily serves.

\subsection{Why Existing One-shot Methods Fail in CUL}
\label{sec:why_fail}

We now discuss \emph{why}, not merely how, existing unlearning methods, designed for single-step deletion, fail when applied sequentially. We identify three contributing factors, each of which motivates a specific component of our proposed framework.

\subsubsection{Coarse Mapping Ignores Locality}

Most guidance-based unlearning methods (i.e., methods that steer the model's predictions away from the forget concept) employ a \emph{prompt-to-phrase} mapping: all forget prompts are redirected to the same fixed anchor (e.g., an empty string or a generic concept)~\cite{gandikota2023erasing,kumari2023ablating,wu2025erasing,fan2024salun,gandikota2024unified}. 
This is a poor substitute for per-prompt mapping because different prompts containing the same forget concept can differ substantially in scene context, attributes, and composition.
Mapping them all to the same surrogate forces \emph{large} score-prediction displacements, much larger than necessary if each prompt were redirected to a nearby, context-appropriate target.
Under repetition, these oversized distortions accumulate into \textbf{cumulative degradation of the learned score field}, progressively eroding the model's generative fidelity~\cite{thakral2025continual,lee2026continual}.
\emph{Our Locality-Aware Target Selection (Section~\ref{sec:target_selection}) addresses this by choosing the nearest safe mapping target for each forget prompt, minimizing the score-prediction displacement per step.}

\subsubsection{Global Replay Cannot Protect Local Neighbors}

Even when methods include retention regularization (e.g., random replay or broad distillation), the regularization signal is spread uniformly across the concept space.
However, the collateral damage from unlearning, the \emph{ripple effect}, is not uniform: it falls disproportionately on concepts semantically close to the forget concept, because these concepts have similar internal representations and score predictions~\cite{amara2025erasebench,zhang2024unlearncanvas}.
Global replay over-invests on already-safe distant concepts and under-invests on the most vulnerable neighbors.
Under repeated steps, the unprotected local neighborhood degrades fastest, causing \textbf{retention collapse}: related retention accuracy ($RR_{\text{acc}}$, the ability to generate concepts semantically close to the forgotten one) drops sharply even when general retention accuracy ($GR_{\text{acc}}$, generation quality on broadly unrelated concepts) appears relatively stable.
\emph{Our Locality-Aware Replay (Section~\ref{sec:replay}) directly addresses this by identifying and reinforcing the concepts closest to the forget concept in the model's own representation.}

\subsubsection{Text-Space Proximity $\neq$ Model-Space Proximity}
\label{sec:why_model_aware}

Text-space similarity (e.g., CLIP embeddings) doesn't reflect how similarly a diffusion model processes two prompts internally: prompts close in CLIP space may still induce very different noise predictions.
Moreover, each unlearning step alters the model's internal landscape, meaning proximity relationships valid before an update may no longer hold afterward.
In CUL, where per-step precision is critical, such mismatches compound across steps.
Our framework therefore uses score-prediction distance, computed directly in the model's noise-prediction space (Section~\ref{sec:score_distance}), for all target and neighbor selection, and recomputes it at every continual step using the current frozen teacher to track the evolving model landscape.

\vspace{0.5em}
\noindent Together, these three factors (excessive updates from coarse mapping, unprotected local neighborhoods, and proxy-based selection) explain why existing methods collapse after 3--5 sequential steps (see Table~\ref{tab:main_comparison}). Our framework addresses each factor with a corresponding component.

\subsection{Denoising Trajectory as an Interpretive Lens}
\label{sec:trajectory_lens}

To build intuition for why locality-aware design helps, we briefly appeal to the denoising-trajectory viewpoint.
A text-to-image diffusion model can be viewed as a score-field estimator: its noise predictions $\epsilon_\theta(z_t, t, p)$ define a vector field that guides noisy latents $z_t$ toward clean images.
Different text prompts induce different trajectories through this field.
When two prompts produce \emph{similar} noise predictions on shared noisy states, their trajectories are close, and redirecting one toward the other requires only a small perturbation of the score field.
Conversely, redirecting to a \emph{distant} target requires a large field distortion that can inadvertently deflect neighboring trajectories.

This perspective explains the benefit of both our components:
(i)~choosing a mapping target with minimal score-prediction distance keeps the field perturbation small, limiting drift;
(ii)~replaying the concepts whose trajectories are nearest to the forget trajectory directly stabilizes the region of the score field most affected by the unlearning update.

\section{Method: Locality-Aware Continual Unlearning}
\label{sec:method}

\subsection{Overview}
\label{sec:overview}

We frame each continual unlearning step as a multi-objective teacher-student process.
The frozen previous-step model $\epsilon_{\hat{\theta}_{i-1}}$ serves as the teacher; the student $\epsilon_{\theta_i}$ is initialized from it and updated to satisfy three complementary objectives:
\begin{itemize}
    \item \textbf{Unlearn} ($\mathcal{L}_{\text{unlearn}}$): redirect the model's response to forget prompts toward a safe, nearby mapping target, editing \emph{locally and minimally} (addresses coarse-mapping failure, Section~\ref{sec:why_fail}).
    \item \textbf{Local replay} ($\mathcal{L}_{\text{retain}}$): preserve the denoising behavior of the concepts nearest to the forget concept, which are most vulnerable to the ripple effect (addresses neighborhood failure).
    \item \textbf{Parameter regularization} ($\mathcal{L}_{\text{reg}}$): a lightweight $\ell_2$ penalty that limits cumulative drift over many steps (addresses parameter drift across sequential steps):
    $\mathcal{L}_{\text{reg}} = \|\theta_i - \hat{\theta}_{i-1}\|_2^2.$
\end{itemize}
The final objective combines all three:
\begin{align}
\mathcal{L}_{\text{total}} = \lambda_{\text{unlearn}} \mathcal{L}_{\text{unlearn}} + \lambda_{\text{retain}} \mathcal{L}_{\text{retain}} + \lambda_{\text{reg}} \mathcal{L}_{\text{reg}}.
\label{eq:total_loss}
\end{align}

A key design principle unifies the first two components: both use \emph{score-prediction distance} (defined next) to make locality-aware decisions in the model's own representation space.
This coupling is the main design difference from using the ingredients independently.
Target selection alone, including AGE-style adaptive replacement, can make the forget update smaller but does not preserve nearby non-target trajectories during training.
Conversely, global replay or parameter regularization, as used in continual-unlearning backbones, constrains drift but can over-regularize the update while still under-protecting the most affected neighbors.
LACU ties the two decisions together: the forget target and the retain neighborhood are selected by the same frozen teacher and the same score-prediction-distance criterion at each continual step.

\subsection{A Model-Aware Locality Metric: Score-Prediction Distance}
\label{sec:score_distance}

Central to our framework is a single metric that defines ``close'' in the \emph{diffusion model's own terms}.
Given two text prompts $p_a$ and $p_b$, a frozen model $\epsilon_{\hat{\theta}}$, and a set of $K$ shared random noisy-latent/timestep pairs $\{(z_{t_k}^{(k)}, t_k)\}_{k=1}^{K}$, the \textbf{score-prediction distance}\footnote{Strictly speaking, this is the mean squared $\ell_2$ distance; we use ``distance'' for brevity since only relative ranking matters in our application.} is:
\begin{align}
d_{\text{score}}(p_a, p_b; \hat{\theta})
= \frac{1}{K}\sum_{k=1}^{K}
\left\|\epsilon_{\hat{\theta}}(z_{t_k}^{(k)}, t_k, p_a)
      -\epsilon_{\hat{\theta}}(z_{t_k}^{(k)}, t_k, p_b)\right\|_2^2.
\label{eq:score_dist}
\end{align}
Here $K$ denotes the number of sampled noisy states; in practice we use $K_f$ samples for target selection (Section~\ref{sec:target_selection}) and $K_r$ samples for replay ranking (Section~\ref{sec:replay}).
Each noisy latent $z_{t_k}^{(k)}$ is constructed by first sampling a clean latent $z_0^{(k)}$ and then applying the forward diffusion process~\cite{ho2020denoising} at a randomly sampled timestep $t_k$: $z_{t_k}^{(k)} = \sqrt{\bar{\alpha}_{t_k}}\, z_0^{(k)} + \sqrt{1 - \bar{\alpha}_{t_k}}\, \epsilon^{(k)}$, where $\epsilon^{(k)} \sim \mathcal{N}(0, \mathbf{I})$. \\

\noindent\textbf{Why random latents suffice.}
We sample $z_0^{(k)} \sim \mathcal{N}(0, \mathbf{I})$ instead of generating DDIM-inverted latents~\cite{song2021denoising}, which would require a costly multi-step inference pass per prompt.
This is valid because score-prediction distance is used for \emph{relative ranking}: all candidates share the same $(z_{t_k}, t_k)$ pairs, so content-independent bias cancels out.
We also verify empirically that the selections agree with DDIM-latent ranking in the majority of cases (see Supplementary).

This metric measures how differently the model \emph{denoises} the same noisy input under two conditioning signals.
A small distance means the model treats the two prompts similarly; redirecting from $p_a$ to $p_b$ therefore requires only a small perturbation of the learned score field.

\noindent\textbf{Why not CLIP similarity?}
CLIP compares prompts in a text-embedding space that is disjoint from the model's conditioning and denoising pathways.
Two prompts can be close in CLIP space yet induce very different noise predictions, and vice versa.
Because unlearning directly modifies the model's parameters, the metric guiding target and neighbor selection should reflect the model's own noise predictions.
Score-prediction distance satisfies this requirement: it operates on the same noisy latent states and model weights that the unlearning update will change, making it a \emph{model-aware} measure of locality.
We validate this empirically in Section~\ref{sec:experiments}, where score-prediction-distance-based selection outperforms CLIP-based ranking for both target selection and replay neighbor selection.

\begin{figure*}[t]
    \centering
    \includegraphics[width=\textwidth]{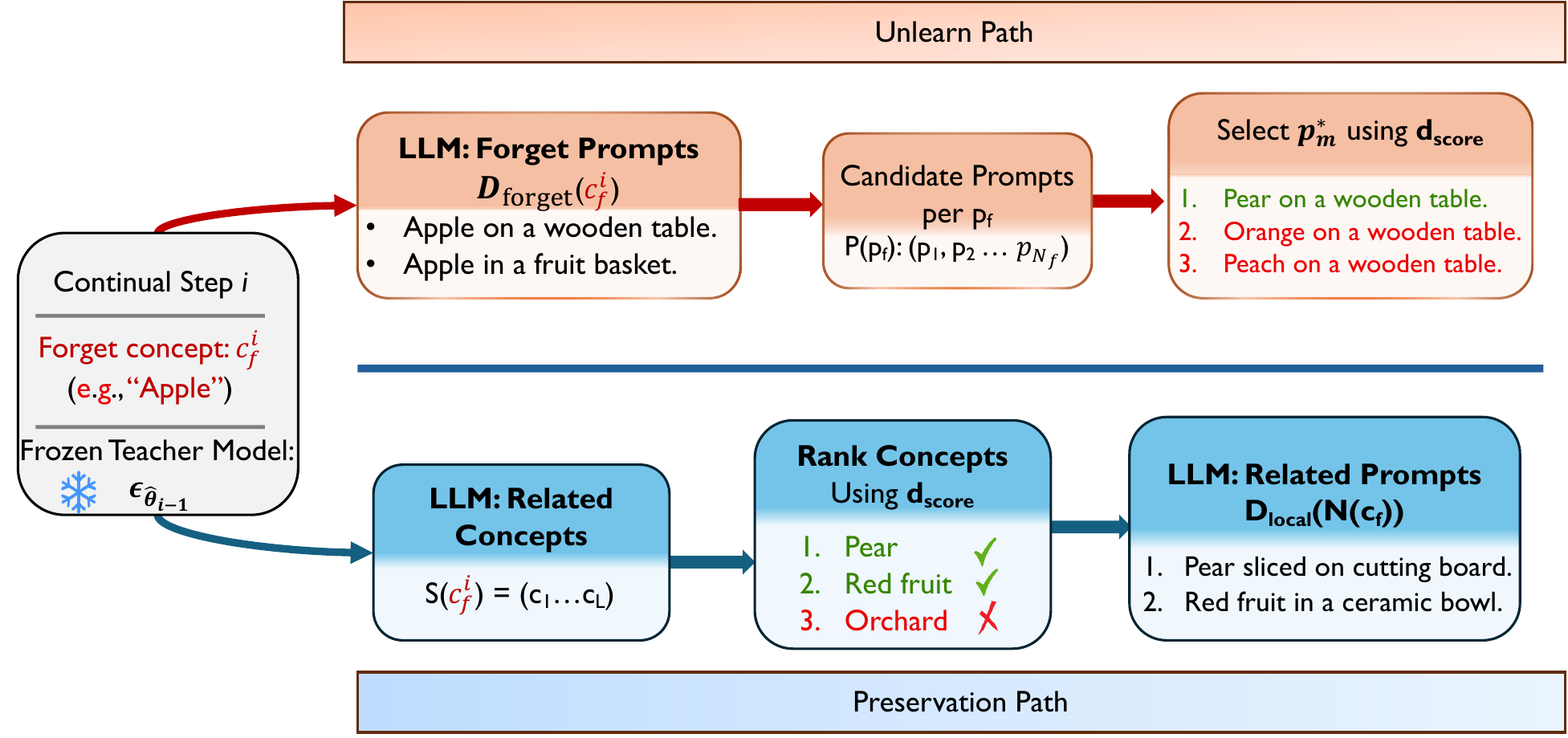}
    \caption{\textbf{Prompt generation and selection pipeline in LACU.}
    At each continual step, the forget concept $c_{\text{f}}^{(i)}$ and the frozen teacher $\epsilon_{\hat{\theta}_{i-1}}$ drive two parallel paths.
    \textbf{Unlearn Path (top):} An LLM generates forget prompts and context-preserving mapping candidates per prompt; the optimal target $p_{\text{m}}^{*}$ is selected via the smallest $d_{\text{score}}$ (Eq.~\ref{eq:traj_align}).
    \textbf{Preservation Path (bottom):} LLM-proposed related concepts are ranked by $d_{\text{score}}$; the top-$N_r$ nearest form the local neighborhood $\mathcal{N}(c_{\text{f}})$ (\textcolor{green!60!black}{\checkmark}), distant ones are discarded (\textcolor{red}{\ding{55}}), and replay prompts are generated for the selected neighbors.}
    \label{fig:data_flow}
\end{figure*}

\subsection{Prompt Design and Data Flow}
\label{sec:prompt_design}

At continual step $i$, $c_{\text{f}}^{(i)}$ denotes the \textbf{forget concept}. A \textbf{forget prompt} $p_{\text{f}}$ instantiates it in a specific scene context; a \textbf{mapping target} $p_{\text{m}}^{*}$ is the safe replacement selected for $p_{\text{f}}$; and a \textbf{replay prompt} $p_{\text{r}}$ belongs to a concept in the local neighborhood of $c_{\text{f}}^{(i)}$.
The complete data flow is illustrated in Fig.~\ref{fig:data_flow}: for each step we generate forget prompts, select per-prompt mapping targets via $d_{\text{score}}$ (Eq.~\ref{eq:traj_align}), identify the nearest retain concepts as replay neighbors via Eq.~\ref{eq:replay_traj}, and alternate between unlearning and replay distillation with $\ell_2$ regularization. LLM prompt templates and selection variants are detailed in Section~\ref{sec:results} and Supplementary Section~8.

\subsection{Component A: Locality-Aware Target Selection}
\label{sec:target_selection}

This component addresses the coarse-mapping failure identified in Section~\ref{sec:why_fail}: instead of redirecting all forget prompts to a single fixed anchor, we select a separate, context-preserving \textbf{mapping target} for each forget prompt via \emph{prompt-to-prompt} mapping.

\noindent\textbf{Example.}
When forgetting Apple, \textit{``an Apple logo engraved on a silver laptop lid on a desk''} is best mapped to \textit{``a tech company logo engraved on a silver laptop lid on a desk,''} whereas \textit{``a sliced apple on a wooden cutting board in kitchen light''} is best mapped to \textit{``a sliced red fruit on a wooden cutting board in kitchen light.''} Although both contain the same concept name, their surrounding context differs, so prompt-level selection is more faithful than a single concept-level surrogate.

\subsubsection{Two-Stage Procedure}
Selection is recomputed at every continual step using the frozen teacher $\epsilon_{\hat{\theta}_{i-1}}$, because the model's denoising landscape changes after each update; targets chosen from an earlier model can become misaligned with the current score field.

\noindent\textbf{Stage 1: Candidate generation.}
An LLM proposes $|\mathcal{P}(p_{\text{f}})|{=}M$ candidate replacement prompts that preserve the non-target scene content of $p_{\text{f}}$ while removing the forget concept.

\noindent\textbf{Stage 2: Model-aware selection.}
We select the candidate with the minimum score-prediction distance to the forget prompt (Eq.~\ref{eq:score_dist}):
\begin{align}
p_{\text{m}}^{*}(p_{\text{f}})=\arg\min_{p\in\mathcal{P}(p_{\text{f}})} \;
d_{\text{score}}(p_{\text{f}},\, p;\; \hat{\theta}_{i-1}).
\label{eq:traj_align}
\end{align}
This yields the mapping target whose score predictions are closest to the forget prompt's, so the required score-prediction displacement is minimized.

\begin{algorithm}[H]
\caption{$\mathcal{L}_{\text{unlearn}}$: Locality-Aware Target Selection}
\label{alg:unlearn}
\begin{algorithmic}[1]
\Require Student $\epsilon_{\theta_i}$, Teacher $\epsilon_{\hat{\theta}_{i-1}}$, forget prompts $\mathcal{D}_{\text{forget}}$, candidate prompt generator $\mathcal{P}(\cdot)$, number of candidates $M$, sample count $K_f$
\Statex \textit{\textbf{// Stage 1: Offline candidate selection (one-time, before training)}}
\For{each $p_{\text{f}} \in \mathcal{D}_{\text{forget}}$}
\State Generate candidates $\mathcal{P}(p_{\text{f}})=\{p_1,\ldots,p_{M}\}$
\State $p_{\text{m}}^{*}(p_{\text{f}}) \leftarrow \arg\min_{p\in\mathcal{P}(p_{\text{f}})} d_{\text{score}}(p_{\text{f}},\, p;\; \hat{\theta}_{i-1})$ \hfill \textit{// Eq.~\ref{eq:traj_align}}
\EndFor
\Statex \textit{\textbf{// Stage 2: Online training (per-iteration optimization)}}
\For{each training iteration}
\State Sample $p_{\text{f}} \sim \mathcal{D}_{\text{forget}}$; look up $p_{\text{m}}^{*}(p_{\text{f}})$
\State $z_0^{\text{u}} \sim \text{DDIM}(\epsilon_{\hat{\theta}_{i-1}}, p_{\text{f}})$
\State Sample $t \sim \mathcal{U}(1, T)$, $\epsilon^{\text{u}} \sim \mathcal{N}(0, \mathbf{I})$
\State $z_{t}^{\text{u}} \leftarrow \sqrt{\bar{\alpha}_{t}}z_0^{\text{u}} + \sqrt{1-\bar{\alpha}_{t}}\epsilon^{\text{u}}$
\State $\epsilon_{\text{student}}^{\text{u}} \leftarrow \epsilon_{\theta_i}(z_{t}^{\text{u}}, t, p_{\text{f}})$\;, \quad \quad \quad $\epsilon_{\text{teacher}}^{\text{u}} \leftarrow \epsilon_{\hat{\theta}_{i-1}}(z_{t}^{\text{u}}, t, p_{\text{m}}^{*}(p_{\text{f}}))$
\State $\mathcal{L}_{\text{unlearn}} \leftarrow \|\epsilon_{\text{teacher}}^{\text{u}} - \epsilon_{\text{student}}^{\text{u}}\|_2^2$
\EndFor
\end{algorithmic}
\end{algorithm}

\subsection{Component B: Locality-Aware Replay}
\label{sec:replay}

This component addresses the unprotected-neighborhood failure in Section~\ref{sec:why_fail}: unlearning perturbs denoising behavior of nearby concepts most strongly, so we focus replay on the \emph{local} neighborhood rather than distributing it globally.

\subsubsection{Procedure}
For each forget concept $c_{\text{f}}^{(i)}$, we first obtain a candidate pool $\mathcal{S}(c_{\text{f}}^{(i)}) = \{c_1, \ldots, c_L\}$ via an LLM that enumerates semantically related concepts (the LLM provides breadth; model-aware ranking provides precision).
We then rank each candidate concept $c_l$ ($l = 1, \ldots, L$) by its score-prediction distance to the forget concept, using the concept names directly as conditioning prompts:
\begin{align}
d_l
= d_{\text{score}}\!\big(c_{\text{f}},\, c_l;\; \hat{\theta}_{i-1}\big),
\label{eq:replay_traj}
\end{align}
where $d_{\text{score}}$ is computed with $K_r$ sampled noisy states (Eq.~\ref{eq:score_dist}, using $K_r$ in place of $K$).
We keep the $N_r$ nearest concepts:
\begin{align}
\mathcal{N}(c_{\text{f}})=\underset{c_l \in \mathcal{S}(c_{\text{f}})}{\operatorname{Top\text{-}N_r}}\big(-d_l\big),
\end{align}
and generate replay prompts from $\mathcal{N}(c_{\text{f}})$ for local retention distillation.

\begin{algorithm}[H]
\caption{$\mathcal{L}_{\text{retain}}$: Locality-Aware Replay}
\label{alg:retain}
\begin{algorithmic}[1]
\Require Student $\epsilon_{\theta_i}$, Teacher $\epsilon_{\hat{\theta}_{i-1}}$, forget concept $c_{\text{f}}$, candidate pool $\mathcal{S}(c_{\text{f}}) = \{c_1, \ldots, c_L\}$, sample count $K_r$, neighbor count $N_r$
\Statex \textit{\textbf{// Stage 1: Offline neighbor selection (one-time, before training)}}
\For{each $c_l \in \mathcal{S}(c_{\text{f}})$}
\State $d_l \leftarrow d_{\text{score}}\!\big(c_{\text{f}},\, c_l;\; \hat{\theta}_{i-1}\big)$ \hfill \textit{// Eq.~\ref{eq:score_dist}}
\EndFor
\State $\mathcal{N}(c_{\text{f}}) \leftarrow \underset{c_l\in\mathcal{S}(c_{\text{f}})}{\operatorname{Top\text{-}N_r}}(-d_l)$
\State Construct local replay prompt set $\mathcal{D}_{\text{local}}(\mathcal{N}(c_{\text{f}}))$
\Statex \textit{\textbf{// Stage 2: Online training (per-iteration optimization)}}
\For{each training iteration}
\State Sample $p_{\text{r}} \sim \mathcal{D}_{\text{local}}(\mathcal{N}(c_{\text{f}}))$
\State $z_0^{\text{r}} \sim \text{DDIM}(\epsilon_{\hat{\theta}_{i-1}}, p_{\text{r}})$
\State Sample $s \sim \mathcal{U}(1, T)$, $\epsilon^{\text{r}} \sim \mathcal{N}(0, \mathbf{I})$
\State $z_{s}^{\text{r}} \leftarrow \sqrt{\bar{\alpha}_{s}}z_0^{\text{r}} + \sqrt{1-\bar{\alpha}_{s}}\epsilon^{\text{r}}$
\State $\epsilon_{\text{student}}^{\text{r}} \leftarrow \epsilon_{\theta_i}(z_{s}^{\text{r}}, s, p_{\text{r}})$\;, \quad \quad \quad $\epsilon_{\text{teacher}}^{\text{r}} \leftarrow \epsilon_{\hat{\theta}_{i-1}}(z_{s}^{\text{r}}, s, p_{\text{r}})$
\State $\mathcal{L}_{\text{retain}} \leftarrow \|\epsilon_{\text{teacher}}^{\text{r}} - \epsilon_{\text{student}}^{\text{r}}\|_2^2$
\EndFor
\end{algorithmic}
\end{algorithm}

\section{Experiments}
\label{sec:experiments}

\begin{table*}[t]
    \centering
    \caption{Quantitative comparison of LACU against existing unlearning methods under continual unlearning. All methods start from the same Stable Diffusion v1.5 checkpoint (\textbf{SD v1.5 base: Accuracy 89\%, CLIP score 32.6}) and unlearn the same 10 concepts; results are averaged over 5 random orderings to eliminate order dependence. Baselines that achieve high $U_{\text{acc}}$ (e.g., EraseFlow, ANT, UCE, AGE) suffer severe retention collapse ($RR_{\text{acc}}/GR_{\text{acc}} \to 0$) over sequential steps, while methods that preserve retention (e.g., DUGE, Sculpt.\ Mem) sacrifice unlearning accuracy. \textbf{LACU balances both retention and unlearning.}}
    \label{tab:main_comparison}
    \small
    \setlength{\tabcolsep}{4pt}
    \resizebox{\textwidth}{!}{
    \begin{tabular}{l | c c c c | c c c c | c c c c}
    \toprule
    \multirow{2}{*}{\textbf{Method}} &
    \multicolumn{4}{c|}{\textbf{After 1 Unlearning}} &
    \multicolumn{4}{c|}{\textbf{After 5 Unlearning}} &
    \multicolumn{4}{c}{\textbf{After 10 Unlearning}} \\
    & $U_{\text{acc}} \uparrow$ & $U_{\text{clip}} \uparrow$ & $RR_{\text{acc}} \uparrow$ & $GR_{\text{acc}} \uparrow$
    & $U_{\text{acc}} \uparrow$ & $U_{\text{clip}} \uparrow$ & $RR_{\text{acc}} \uparrow$ & $GR_{\text{acc}} \uparrow$
    & $U_{\text{acc}} \uparrow$ & $U_{\text{clip}} \uparrow$ & $RR_{\text{acc}} \uparrow$ & $GR_{\text{acc}} \uparrow$ \\
    \midrule
    ESD-u~\cite{gandikota2023erasing}     &                      0.97 &                      29.1 &                      0.60 &                      0.75 &                      0.83 &                      24.7 &                      0.30 &                      0.35 &                      0.93 &                      21.3 &                      0.13 &                      0.15 \\
    ESD-x~\cite{gandikota2023erasing}     &  \cellcolor{tabfirst}0.99 &                      27.3 &                      0.55 &                      0.77 &                      0.93 &                      23.0 &                      0.29 &                      0.28 &                      0.96 &                      19.0 &                      0.09 &                      0.08 \\
    UCE~\cite{gandikota2024unified}       &  \cellcolor{tabfirst}0.99 &                      21.2 &                      0.65 &                      0.39 &                      0.86 &                      21.5 &                      0.30 &                      0.26 &                      0.96 &                      19.5 &                      0.08 &                      0.05 \\
    MACE~\cite{lu2024mace}                &  \cellcolor{tabfirst}0.99 &                      28.2 &                      0.69 &                      0.83 &                      0.94 &                      18.5 &                      0.06 &                      0.03 & \cellcolor{tabthird}0.97 &                      19.4 &                      0.03 &                      0.02 \\
    Meta~\cite{gao2025meta}               & \cellcolor{tabthird}0.97 &                      27.76 &                      0.45 &                      0.78 & \cellcolor{tabthird}0.95 &                      20.1 &                      0.10 &                      0.09 &  \cellcolor{tabfirst}1.00 &                      18.2 &                      0.02 &                      0.01 \\
    EraseFlow~\cite{kusumba2025eraseflow} & \cellcolor{tabsecond}0.98 &                      23.9 &                      0.27 &                      0.53 &                      0.92 &                      18.0 &                      0.07 &                      0.02 & \cellcolor{tabthird}0.97 &                      19.4 &                      0.03 &                      0.02 \\
    ANT~\cite{li2025set}                  & \cellcolor{tabsecond}0.98 &                      27.2 &                      0.49 &                      0.75 &  \cellcolor{tabfirst}1.00 &                      18.1 &                      0.02 &                      0.01 &  \cellcolor{tabfirst}1.00 &                      20.5 &                      0.00 &                      0.00 \\
    CA~\cite{kumari2023ablating}          &                      0.95 &                      28.1 & \cellcolor{tabsecond}0.73 & \cellcolor{tabthird}0.84 &                      0.89 &                      27.5 &                      0.58 &                      0.64 &                      0.90 &                      22.8 &                      0.34 &                      0.34 \\
    AGE~\cite{bui2025fantastic}           &  \cellcolor{tabfirst}0.99 &                      25.2 &                      0.49 &                      0.82 & \cellcolor{tabsecond}0.99 &                      21.4 &                      0.36 &                      0.55 & \cellcolor{tabsecond}0.99 &                      19.1 &                      0.15 &                      0.23 \\
    DUGE~\cite{thakral2025continual}      &                      0.96 &  \cellcolor{tabthird}29.7 &  \cellcolor{tabthird}0.70 & \cellcolor{tabsecond}0.85 &                      0.69 &  \cellcolor{tabfirst}32.6 & \cellcolor{tabsecond}0.73 & \cellcolor{tabsecond}0.80 &                      0.74 &  \cellcolor{tabfirst}30.7 & \cellcolor{tabsecond}0.62 & \cellcolor{tabsecond}0.73 \\
    SMem~\cite{li2025sculpting}           &                      0.96 &  \cellcolor{tabfirst}31.4 &  \cellcolor{tabthird}0.70 &  \cellcolor{tabthird}0.84 &                      0.64 & \cellcolor{tabsecond}31.5 &  \cellcolor{tabthird}0.72 &  \cellcolor{tabthird}0.78 &                      0.73 & \cellcolor{tabsecond}29.3 &                      0.58 &                      0.67 \\
    Grad Proj~\cite{lee2026continual}     &  \cellcolor{tabfirst}0.99 &                      26.3 &                      0.68 & \cellcolor{tabsecond}0.85 &                      0.86 &                      28.7 &                      0.69 &  \cellcolor{tabthird}0.78 &                      0.83 &                      28.1 &  \cellcolor{tabthird}0.61 &  \cellcolor{tabthird}0.71 \\
    \midrule
    \textbf{LACU (Ours)}                  &  \cellcolor{tabfirst}0.99 & \cellcolor{tabsecond}30.0 &  \cellcolor{tabfirst}0.84 &  \cellcolor{tabfirst}0.89 &                      0.89 &  \cellcolor{tabthird}29.7 &  \cellcolor{tabfirst}0.80 &  \cellcolor{tabfirst}0.87 &                      0.90 &  \cellcolor{tabthird}29.1 &  \cellcolor{tabfirst}0.72 &  \cellcolor{tabfirst}0.87 \\

    \bottomrule
    \end{tabular}
    }
\end{table*}

We evaluate whether locality-aware design, using score-prediction distance for both target selection and replay, improves sequential unlearning stability. Specifically, we test two hypotheses: (i) locality-aware target selection reduces cumulative drift, and (ii) locality-aware replay improves neighborhood retention. This section details the setup and metrics used to test both.

\subsection{Experimental Settings}

\noindent\textbf{Base Model and Training.}
All experiments are performed on the Stable Diffusion v1.5 model~\cite{rombach2022high}, which is the commonly used base model for experimentation (see Supplementary for other SD models and full training details). Each unlearning step is trained for 250--350 optimizer steps using AdamW on 3 NVIDIA H100 GPUs across the 10-concept sequence; each unlearning update takes approximately 60 minutes.
Timesteps $t$ and $s$ are sampled uniformly from 0 to 600.

\vspace{\baselineskip}
\noindent\textbf{Prompt Generation and Mapping.}
To construct the datasets for unlearning and preservation, we use an automated prompt-construction pipeline to generate diverse and semantically rich prompts.
\begin{itemize}
    \item \textbf{Forget and Candidate Mapping Prompts:} For each concept targeted for unlearning, we generated $N_f{=}100$ unique forget prompts ($\mathcal{D}_{\text{forget}}$). For each forget prompt, we construct a candidate pool of $M{=}10$ context-preserving mapping prompts. The LLM prompt templates explicitly instruct exclusion of the forget concept while preserving context. We then select the final prompt-level mapping target using Eq.~\ref{eq:traj_align} with $K_f$ scoring samples.
    \item \textbf{Retain Prompts:} For each forget concept, we obtain related candidate concepts via an LLM, then rank their proximity in noise-prediction space using Eq.~\ref{eq:replay_traj} with $K_r$ scoring samples. We keep the top-$N_r{=}10$ nearest concepts, generate replay prompts from them, and mix these with random global prompts for broader coverage. In ablations, we compare this score-prediction-distance-based ranking against CLIP-based neighbor ranking.
\end{itemize}
The LLM is used only to populate candidate pools; the LACU objective itself is agnostic to how these candidates are produced. In settings where LLM use is undesirable, the same score-prediction-distance selection can operate over templates, curated vocabularies, or human-provided candidate prompts and related concepts. In a no-LLM variant, we replace the candidate generator with template-based forget/mapping prompts and a same-category pool of related concepts. Its 10-step $U_{\text{acc}}$, $RR_{\text{acc}}$, and $GR_{\text{acc}}$ are 0.85, 0.81, and 0.87, respectively, indicating that model-aware scoring and local replay drive the retention stability.

\vspace{\baselineskip}
\noindent\textbf{Unlearning Targets.}
We evaluated our method by sequentially unlearning a diverse set of 10 concepts, comprising objects, artistic styles, and famous individuals: \textit{Pikachu, Brad Pitt, Golf Ball, Van Gogh Style, Apple, Spiderman, Lionel Messi, Cartoon Style, Banana} and \textit{Mickey Mouse}. To eliminate dependence on a particular deletion order, we run each method over 5 different random permutations of these 10 concepts and report the average metrics.

\subsection{Evaluation Metrics}

We evaluate across three axes: unlearning efficacy, neighborhood retention, and global retention. Our evaluation extends UnlearnCanvas and EraseBench protocols~\cite{zhang2024unlearncanvas,amara2025erasebench}.

\vspace{\baselineskip}
\noindent\textbf{Unlearning and Retention Accuracy.}
To automate and standardize evaluation, we use the Qwen2.5-VL-7B-Instruct model~\cite{bai2025qwen2} as a VLM judge, querying it with binary questions on generated images. The VLM is used only for evaluation; target and replay selection are decided by score-prediction distance in the diffusion model's prediction space. VLM-based evaluation has become standard practice in recent T2I unlearning studies, with SEE, EraseBench, CA, SA, and UnGuide using VLMs as classifiers or judges~\cite{saha-etal-2025-side,amara2025erasebench,kumari2023ablating,heng2023selective,polowczyk2025unguide}. We found VLMs to be substantially more reliable than ImageNet-style classifiers or UnlearnCanvas heads, which struggle with diverse concepts (celebrities, artistic styles, etc.). A judge-sensitivity check with alternative VLMs changed the key rows by at most $\pm 2\%$ and preserved the relative ranking of LACU and the baselines. All evaluation prompts are unseen during training.
\begin{itemize}
    \item \textbf{Unlearning Accuracy ($U_{\text{acc}}$) \& CLIP Score ($U_{\text{clip}}$):} $U_{\text{acc}}$ measures forgetting effectiveness. $U_{\text{clip}}$ measures ``in-prompt retainability''~\cite{ren2025sixcd}, ensuring benign parts of the prompt are still generated correctly. The ideal outcome is both high $U_{\text{acc}}$ and high $U_{\text{clip}}$.
    \item \textbf{Related Retention Accuracy ($RR_{\text{acc}}$) \& CLIP Score ($RR_{\text{clip}}$):} $RR_{\text{acc}}$ measures the unintended impact on semantically adjacent concepts~\cite{amara2025erasebench,zhang2024unlearncanvas} (e.g., testing ``Impressionism'' after unlearning ``Van Gogh''). $RR_{\text{clip}}$ ensures text-to-image alignment for these concepts is preserved.
    \item \textbf{General Retention Accuracy ($GR_{\text{acc}}$) \& CLIP Score ($GR_{\text{clip}}$):} $GR_{\text{acc}}$ assesses overall knowledge preservation on general unrelated prompts; $GR_{\text{clip}}$ verifies text-to-image alignment remains intact.
\end{itemize}
The main tables report the compact accuracy view, while the Supplementary reports the full CLIP-score breakdown ($U_{\text{clip}}$, $RR_{\text{clip}}$, and $GR_{\text{clip}}$) for all steps and methods.
 
\noindent\textbf{Generation Quality:} We adopt Fréchet Inception Distance (FID) to evaluate image quality and diversity, following UnlearnCanvas benchmark methodology.

\subsection{Experimental Results and Analysis}

\begin{figure*}[t]
    \centering
    \includegraphics[width=\textwidth]{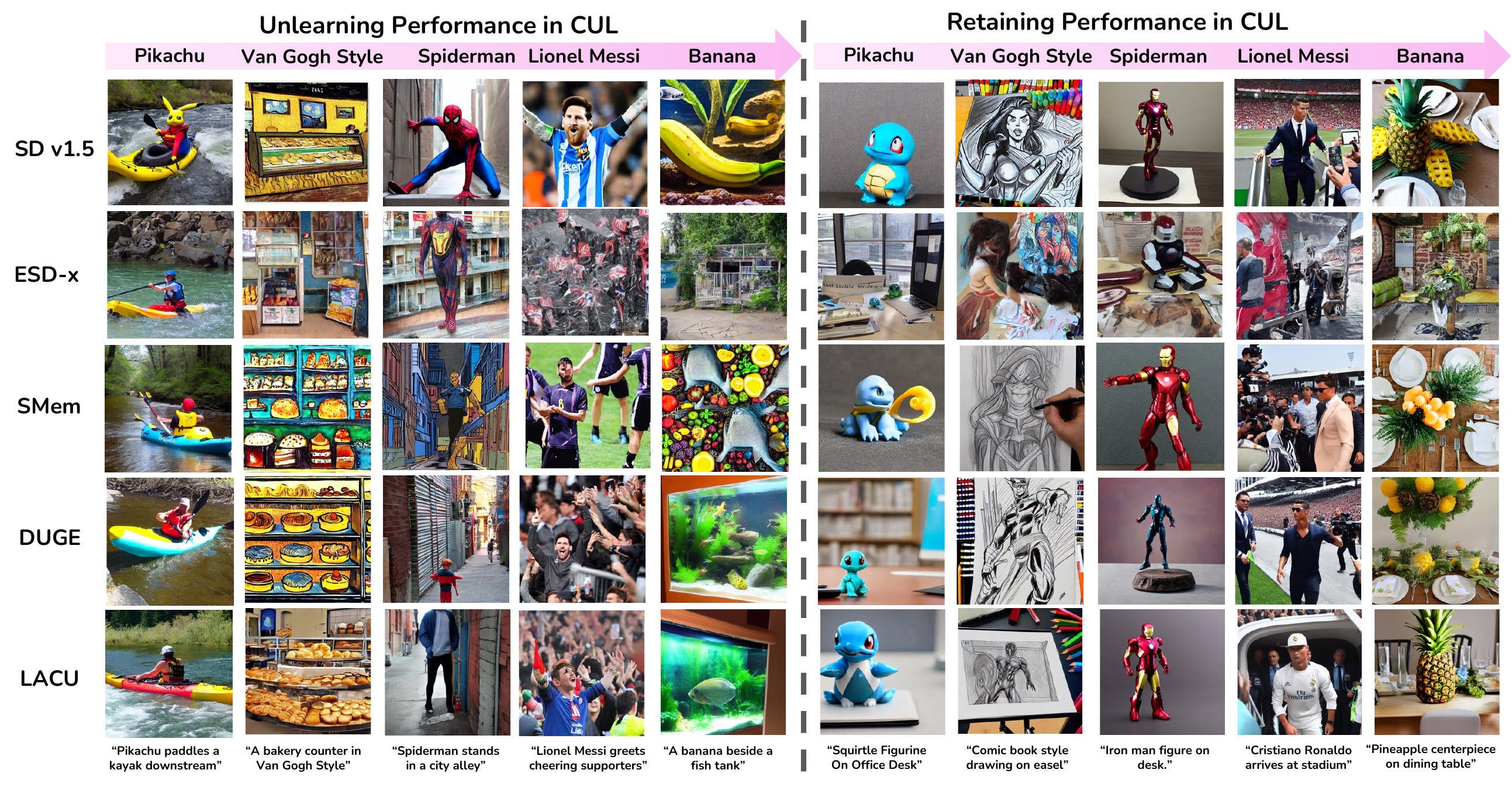}
    \captionof{figure}{Qualitative comparison after 10 sequential unlearning steps. Baselines like ESD-x~\cite{gandikota2023erasing} suffer catastrophic retention collapse on retained concepts (right), while DUGE~\cite{thakral2025continual} and SMem~\cite{li2025sculpting} avoid total collapse but still degrade. Our method (bottom row) effectively unlearns target concepts (left) while preserving generative quality on retained concepts (right).}
    \label{fig:teaser}
\end{figure*}

We have conducted extensive evaluations to assess the effectiveness of our locality-aware framework in the continual unlearning setup. \\

\textbf{How well do existing methods perform under sequential requests?} 
Across 10 sequential steps, existing methods exhibit a clear stability--efficacy trade-off. Aggressive methods (ESD, ANT, EraseFlow) show near-perfect unlearning but destroy retention via compounding parameter drift. Conservative methods (DUGE, SMem) preserve retention but sacrifice unlearning efficacy. This confirms that one-shot objectives do not transfer to sequential operation. \\
Most one-shot baselines use fixed anchors or mapping concepts, so repeated updates accumulate coarse redirections. AGE~\cite{bui2025fantastic} improves target choice with an adaptive target, but Table~\ref{tab:main_comparison} shows that target selection alone still performs poorly in CUL because nearby non-target score trajectories remain unprotected. \\

\textbf{Does our framework improve continual unlearning?}
Yes. Our results show consistent gains across all core criteria. In $GR_{\text{acc}}/RR_{\text{acc}}$ trend curves over 10 steps, our method remains substantially flatter and higher than baselines, indicating stronger stability over sequential steps. These trends support the claim that locality-aware target selection and replay reduce cumulative distortion under repeated unlearning. \\
The dynamic nature of LACU also avoids a failure case of fixed-anchor methods. If a concept previously used as a mapping target later becomes a forget concept, a fixed mapping can degenerate into an ineffective self-mapping. LACU instead re-scores candidate trajectories with the current frozen teacher at each step and selects a new nearest safe target, so the model can still remove the newly requested concept; in our runs this produced no degradation in $U_{\text{acc}}$, with only slightly slower convergence. \\

\begin{table*}[t]
    \centering
    \caption{Comparison of mapping-target and replay-neighbor selection variants after 1, 5, and 10 unlearning steps. Naive/far mappings deteriorate rapidly, while score-prediction-distance-based selection (bottom row) provides the strongest long-term balance, notably the highest $RR_{\text{acc}}$ and $GR_{\text{acc}}$ at 10 steps.}
    \label{tab:prompt_variants}
    \small
    \setlength{\tabcolsep}{4pt}
    \resizebox{\textwidth}{!}{
    \begin{tabular}{l | c c c c | c c c c | c c c c}
    \toprule
    \multirow{2}{*}{\textbf{Method}} &
    \multicolumn{4}{c}{\textbf{After 1 Unlearning Step}} &
    \multicolumn{4}{c}{\textbf{After 5 Unlearning Steps}} &
    \multicolumn{4}{c}{\textbf{After 10 Unlearning Steps}} \\
    & $U_{\text{acc}} \uparrow$ & $U_{\text{clip}} \uparrow$ & $RR_{\text{acc}} \uparrow$ & $GR_{\text{acc}} \uparrow$
    & $U_{\text{acc}} \uparrow$ & $U_{\text{clip}} \uparrow$ & $RR_{\text{acc}} \uparrow$ & $GR_{\text{acc}} \uparrow$
    & $U_{\text{acc}} \uparrow$ & $U_{\text{clip}} \uparrow$ & $RR_{\text{acc}} \uparrow$ & $GR_{\text{acc}} \uparrow$ \\
    \midrule
    LACU (naive)
    & 0.89 & 24.2 & \cellcolor{tabfirst}0.88 & \cellcolor{tabsecond}0.88 & 0.84 & 22.6 & 0.57 & 0.57 & 0.86 & 22.9 & 0.41 & 0.55 \\

    LACU (LLM)
    & \cellcolor{tabsecond}0.99 & \cellcolor{tabthird}28.7 & \cellcolor{tabsecond}0.86 & \cellcolor{tabthird}0.87 & \cellcolor{tabfirst}0.92 & \cellcolor{tabthird}29.5 & \cellcolor{tabsecond}0.79 & \cellcolor{tabthird}0.80 & \cellcolor{tabfirst}0.95 & \cellcolor{tabthird}28.1 & \cellcolor{tabthird}0.60 & \cellcolor{tabthird}0.76 \\

    LACU (CLIP)
    & \cellcolor{tabthird}0.94 & \cellcolor{tabfirst}30.3 & 0.81 & \cellcolor{tabthird}0.87 & \cellcolor{tabthird}0.88 & \cellcolor{tabsecond}29.6 & \cellcolor{tabthird}0.77 & \cellcolor{tabsecond}0.81 & \cellcolor{tabthird}0.90 & \cellcolor{tabsecond}28.2 & \cellcolor{tabsecond}0.61 & \cellcolor{tabsecond}0.78 \\

    LACU (score)
    & \cellcolor{tabfirst}1.00 & \cellcolor{tabsecond}30.0 & \cellcolor{tabthird}0.84 & \cellcolor{tabfirst}0.89 & \cellcolor{tabsecond}0.90 & \cellcolor{tabfirst}29.8 & \cellcolor{tabfirst}0.81 & \cellcolor{tabfirst}0.87 & \cellcolor{tabsecond}0.92 & \cellcolor{tabfirst}29.2 & \cellcolor{tabfirst}0.74 & \cellcolor{tabfirst}0.87 \\

    \bottomrule
    \end{tabular}
    }
\end{table*}

\textbf{How does the choice of mapping target and replay neighbors affect continual unlearning?}
\label{sec:results}
Table~\ref{tab:prompt_variants} compares four selection strategies within the same LACU pipeline: naive/fixed anchors, direct LLM replacements, CLIP-ranked, and score-prediction-distance-ranked (ours). Naive mappings degrade fastest because a single generic anchor forces large score-prediction displacements that compound over steps. CLIP-ranked variants improve over naive and LLM but still drift as CLIP similarity is decoupled from the model's denoising dynamics. Score-prediction-distance selection achieves the strongest long-term balance because it directly measures how differently the model processes two prompts, ensuring each update is minimally disruptive. This confirms that \emph{model-aware} selection is essential for stability under repeated unlearning. \\

\begin{wraptable}[11]{l}{0.49\textwidth}
    \centering
    \small
    \vspace{-25pt}
    \caption{Ablation after 10 CUL steps. Replay and regularization together are needed to balance unlearning and retention.}
    \label{tab:ablation}
    \setlength{\tabcolsep}{3pt}
    \renewcommand{\arraystretch}{1.04}
    \resizebox{\linewidth}{!}{%
    \begin{tabular}{l c c c c}
        \toprule
        \textbf{Method} & $U_{\text{acc}}$ & $U_{\text{clip}}$ & $RR_{\text{acc}}$ & $GR_{\text{acc}}$ \\
        \midrule
        $\mathcal{L}_{\text{unl}}$ only & \cellcolor{tabfirst}0.94 & 27.2 & 0.32 & 0.67 \\
        $\mathcal{L}_{\text{unl}}{+}\mathcal{L}_{\text{ret}}$ & \cellcolor{tabthird}0.91 & \cellcolor{tabsecond}28.6 & \cellcolor{tabsecond}0.61 & \cellcolor{tabsecond}0.85 \\
        $\mathcal{L}_{\text{unl}}{+}\mathcal{L}_{\text{reg}}$ & \cellcolor{tabsecond}0.92 & \cellcolor{tabthird}28.0 & \cellcolor{tabthird}0.39 & \cellcolor{tabthird}0.78 \\
        \textbf{LACU} & \cellcolor{tabsecond}0.92 & \cellcolor{tabfirst}29.2 & \cellcolor{tabfirst}0.74 & \cellcolor{tabfirst}0.87 \\
        \bottomrule
    \end{tabular}
    }
\end{wraptable}

\textbf{How does our method achieve this, and what is the contribution of each objective?}
Table~\ref{tab:ablation} ablates each component after 10 CUL steps, with all variants using score-prediction-distance-based selection.
$\mathcal{L}_{\text{unlearn}}$ \textbf{alone} forgets effectively but suffers retention collapse from compounding ripple effects.
Adding replay ($\mathcal{L}_{\text{retain}}$) directly counters this by forcing the student to mimic the teacher on nearby concepts, providing functional stability in the local neighborhood.
Adding only regularization ($\mathcal{L}_{\text{reg}}$) prevents cumulative parameter drift but provides weaker retention recovery since it does not explicitly protect vulnerable neighbors.
The \textbf{full LACU} combines both: replay shields the local neighborhood while regularization constrains long-term parameter drift, achieving the best overall balance across all metrics.

\section{Conclusion}
We presented Locality-Aware Continual Unlearning (LACU), a framework that stabilizes sequential concept removal in diffusion models by grounding both target selection and replay in a single model-aware principle: score-prediction distance. Locality-Aware Target Selection chooses the nearest safe mapping target for each forget prompt, keeping the score-prediction displacement as small as possible and limiting cumulative degradation of the learned score field. Locality-Aware Replay identifies and reinforces the retain concepts closest to the forget concept in the model's noise-prediction space, directly shielding the neighborhood where collateral damage concentrates. Together with teacher-student distillation and $\ell_2$ regularization, these components yield stable 10-step continual unlearning with substantially higher related retention ($RR_{\text{acc}}$) and general retention ($GR_{\text{acc}}$) than existing baselines, demonstrating that locality-aware design is key to stability over sequential steps. \\

\noindent\textbf{Limitations.}
Our implementation uses an LLM to generate candidate prompts and related concepts, although the core score-prediction-distance selection and training objectives do not depend on LLMs. The quality and coverage of the candidate pool, whether generated automatically or curated manually, bounds the effectiveness of downstream model-aware selection.
Local replay also introduces an inherent precision trade-off when the forget concept and retained neighbors share heavily overlapping visual features. LACU mitigates this by regularizing only the nearest neighbor trajectories rather than the full retain distribution, which keeps object and identity unlearning strong, but style concepts remain harder because their features are more broadly shared.
More broadly, current continual unlearning methods, including ours, have been evaluated on a limited number of sequential deletions, whereas real-world deployment may require hundreds or thousands of removals over a model's lifetime; developing methods that scale gracefully to such regimes while maintaining strong retention remains an important open challenge.
Finally, the teacher-student distillation framework requires maintaining a frozen copy of the model alongside the trainable student, increasing memory and computation relative to single-model approaches.

\section*{Acknowledgements}
We sincerely thank Satoshi Hayakawa for providing valuable feedback prior to submission.

\bibliographystyle{splncs04}
\bibliography{main}

\end{document}